\documentclass{article}
\pdfoutput=1
\usepackage{microtype}
\usepackage{graphicx}
\usepackage{soul,xcolor}
\usepackage{subcaption}
\usepackage{makecell}
\usepackage{multirow}
\usepackage{booktabs} 

\usepackage{tabularx}
\usepackage{diagbox}

\definecolor{our_red}{HTML}{c92d39}
\definecolor{our_yellow}{HTML}{EDB732}
\setstcolor{our_red}

\def\ccar#1{%
    \pgfmathsetmacro\calc{(-#1)*100/(40)}%
    \edef\clrmacro{\noexpand\cellcolor{our_red!\calc}}%
    \clrmacro%
    \ifdim \calc pt>50pt\color{white}\fi{#1}%
}

\def\ccay#1{%
    \pgfmathsetmacro\calc{(#1)*100/(2)}%
    \edef\clrmacro{\noexpand\cellcolor{our_yellow!\calc}}%
    \clrmacro%
    \ifdim \calc pt>50pt\color{black}\fi{#1}%
}

\usepackage[accepted]{icml2022}

\usepackage{amsmath}
\usepackage{amssymb}
\usepackage{hyperref}
\usepackage{xspace}
\usepackage{graphicx}
\usepackage{booktabs}
\usepackage[utf8]{inputenc} 
\usepackage[T1]{fontenc}
\usepackage[normalem]{ulem}
\usepackage{tikz,siunitx}
\usepackage{nicefrac}
\usepackage{xcolor,colortbl}
\usepackage{babel}
\tikzstyle{bag} = [align=center]

\icmltitlerunning{Contrastive Learning for Combinatorial Problems}

\begin{document}

\twocolumn[
\icmltitle{Augment with Care: Contrastive Learning for Combinatorial Problems}



\icmlsetsymbol{equal}{*}

\begin{icmlauthorlist}
\icmlauthor{Haonan Duan}{equal,to,vec}
\icmlauthor{Pashootan Vaezipoor}{equal,to,vec}
\icmlauthor{Max B. Paulus}{go}
\icmlauthor{Yangjun Ruan}{to,vec}
\icmlauthor{Chris J. Maddison}{to,vec}
\end{icmlauthorlist}

\icmlaffiliation{to}{University of Toronto}
\icmlaffiliation{vec}{Vector Institute}
\icmlaffiliation{go}{ETH Z\"{u}rich}

\icmlcorrespondingauthor{Haonan Duan}{haonand@cs.toronto.edu}

\icmlkeywords{Machine Learning, ICML}

\vskip 0.3in
]




\printAffiliationsAndNotice{\icmlEqualContribution} 
\newcommand{\ve}{VE\xspace}
\newcommand{\au}{AU\xspace}
\newcommand{\satcr}{CR\xspace}
\newcommand{\satsc}{SC\xspace}
\newcommand{\dc}{DC\xspace}
\newcommand{\dv}{DV\xspace}
\newcommand{\lp}{LP\xspace}
\newcommand{\sg}{SG\xspace}
\newcommand{\ppa}{LPA\xspace}
\newcommand{\paa}{LAA\xspace}
\newcommand{\lpa}{LPA\xspace}
\newcommand{\lpas}{LPAs\xspace}
\newcommand{\laas}{LAAs\xspace}
\newcommand{\supp}{\texttt{supp}\xspace}

\begin{abstract}
Supervised learning can improve the design of state-of-the-art solvers for combinatorial problems, but labelling large numbers of combinatorial instances is often impractical due to exponential worst-case complexity. Inspired by the recent success of contrastive pre-training for images, we conduct a scientific study of the effect of augmentation design on contrastive pre-training for the Boolean satisfiability problem. While typical graph contrastive pre-training uses label-agnostic augmentations, our key insight is that many combinatorial problems have well-studied invariances, which allow for the design of \textit{label-preserving augmentations}. We find that label-preserving augmentations are critical for the success of contrastive pre-training. We show that our representations are able to achieve comparable test accuracy to fully-supervised learning while using only 1\% of the labels. We also demonstrate that our representations are more transferable to larger problems from unseen domains. Our code is available at \url{https://github.com/h4duan/contrastive-sat}. 
\end{abstract}

\section{Introduction}
\label{sec:introduction}

Combinatorial problems, e.g., Boolean satisfiability (SAT) or mixed-integer linear programming (MILP), have many applications in the industry and can encode many fundamental computational tasks. These problems are NP-complete, so solvers that perform efficiently in the worst case are not within reach. However, learning can be used to improve the average complexity of solvers on the population of combinatorial problems found in the wild \citep{nair2020solving}.

Supervised learning is a promising approach to combinatorial solver design \citep[e.g.,][]{selsam2019guiding,nair2020solving}. Unfortunately, the need for labels is a severe limitation. Many read-world problems, such as cryptography SAT instances, are extremely hard or, in some cases, even impossible to solve \citep{nejati2019cdcl}. Computing expert branching labels for large-scale MILPs requires sophisticated parallel solvers \citep{nair2020solving}. 

In order to scale learning for combinatorial problems, we ask: how much can we learn from \emph{unlabelled} combinatorial instances?
In this work, we consider a contrastive learning approach, which begins by creating multiple ``views'' of every unlabelled instance, a process called augmentation. An encoder is trained to maximize the similarity between the representations of augmentations that come from the same instance, while minimizing the similarity between those of distinct ones \citep{chen2020simple}. This has been successful in computer vision: contrastive representations can be used with linear predictors to achieve competitive accuracies on ImageNet using a fraction of the labelled instances \citep{chen2020simple, he2020momentum, chen2020improved}. 

Our key insight is that combinatorial problems have well-studied invariances that can be used to design extremely effective augmentations for contrastive learning. Contrastive learning theory indicates that augmentations should be (roughly) label-preserving in order to confer guarantees on downstream prediction \citep{arora2019theoretical, tosh2021contrastive, haochen2021provable}. This is in contrast to the majority of label-agnostic graph contrastive frameworks \citep[e.g.,][]{you2020graph,hassani2020contrastive}. For some combinatorial problems, label-preserving transformations are available from subroutines of existing solvers, e.g.,  variable elimination modifies SAT formulas while preserving their satisfiability. Our augmentations produce new formulas by randomly applying such satisfiability-preserving transformations. Crucially, our augmentations do not require full solves and are much cheaper to compute than the labels.

We study data augmentations for contrastive learning for Boolean satisfiability. We demonstrate that:
\begin{itemize}
    \itemsep0em 
    \item Augmentation design is critical for contrastive pre-training on combinatorial problems. In particular, our augmentations are sufficient for strong performance, while existing graph augmentations are not.
    \item Our contrastive method matches the best supervised baseline while using $100\times$ fewer training labels.
    \item Our contrastive representations transfer better to new problem types than supervised representations.
\end{itemize}

\section{Background}
\label{sec: background}

\textbf{SAT.} A Boolean formula in propositional logic consists of 
Boolean variables composed by logical operators ``and'' ($\land$), ``or'' ($\lor$) and ``not'' ($\lnot$). A \emph{literal} is a variable $v$ or its negation $\lnot v$. A \emph{clause} is a disjunction of literals $\bigvee_{i=1}^n l_i$.  A Boolean formula is in \emph{Conjunctive Normal Form} (CNF) if it is a conjunction of clauses. We assume that all formulas are in CNF. 
A variable assignment satisfies a clause when at least one of its
literals is satisfied. A formula $\phi$ is satisfied under $\pi$ when all of its clauses are satisfied, and $\pi$ is a 
\emph{satisfying assignment} for $\phi$. \emph{SAT} is the problem of deciding, for a given formula $\phi$, if there exists a satisfying assignment (SAT) or not (UNSAT).

\begin{figure}[tb]
    \centering
    \includegraphics[width=0.6\columnwidth]{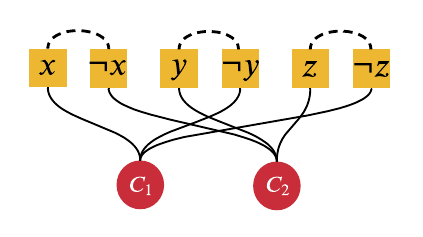}
    \caption{A bipartite graph representation ($\text{LIG}^{+}$) of a SAT formula $\phi:=(x \vee \neg y \vee \neg z) \wedge (\neg x \vee y \vee z)$. The subgraph without \protect\tikz[baseline]{\protect\draw[line width=0.5mm,densely dashed] (0,.8ex)--++(0.5,0);} edges is the LIG of $\phi$.}
    \label{fig:sat_representation}
\end{figure}

We can represent a SAT formula $\phi$ by a bipartite graph 
called \emph{literal-clause incidence graph} (LIG) (Figure \ref{fig:sat_representation}). The graph contains a node for every clause and literal of $\phi$. An edge connects a clause node and a literal node iff the clause contains that literal. We define $\text{LIG}^{+}$ when the literal nodes of the same variable in LIG are connected.

\textbf{NeuroSAT.} \citet{selsam2018learning} proposed NeuroSAT to classify satisfiability of Boolean formulas. NeuroSAT consists of two parts: \textbf{1) \emph{Encoder:}} a special \emph{Graph Neural Network} (GNN) that takes in the $\text{LIG}^{+}$ representation of a formula and produces an embedding for its literals; 
\textbf{2) \emph{Aggregator}:} A function that maps each literal representation to a vote and then aggregates them into a prediction about satisfiability.
More details can be found in Appendix \ref{appendix: neurosat}.

\section{Related Work}
\label{sec:related_work}

\textbf{Graph Contrastive Learning.} Existing graph contrastive frameworks can, in principle, be used to learn representations for combinatorial problems. Common augmentations in graph contrastive frameworks include: 1) perturbing structures, such as, node dropping, subgraph sampling  or graph diffusion, and 2) perturbing features, such as, masking or adding noise to the node features \citep{hassani2020contrastive}. These augmentations have achieved success in multiple graph-level tasks \citep{you2020graph, hassani2020contrastive}, as well as node-level tasks \citep{zhu2020deep,wan2021contrastive,tong2021directed}.

\textbf{Supervised Learning for Combinatorial Optimization (CO).} Almost all modern approaches to machine learning for CO, use variations of a GNN architecture. On the supervised front, the work of \citet{nowak2018revised} on 
\emph{Quadratic Assignment Problem}, \citet{joshi2019efficient} and \citet{prates2019learning} on \emph{Travelling Salesman Problem} (TSP) have shown encouraging results.
\citet{gasse2019exact} trained a branching heuristic for MILPs by imitating an expert policy. Later, \citet{nair2020solving} extended those ideas 
to make them scalable to substantially larger instances. In SAT, \citet{selsam2018learning} proposed a way to train GNNs to solve SAT problems in an end-to-end
fashion. The same architecture was used in \citet{selsam2019guiding} to guide variable branching across conventional solvers.

\textbf{Reinforcement Learning for CO.}
Supervised learning is bottlenecked by the need for labels. Consequently, many have explored the use of \emph{Reinforcement Learning} (RL). Node selection policy of \citet{dai2017learning} for TSP and \citet{kool2018attention} for \emph{Vehicle Routing Problem} are examples of that effort. \citet{LedermanRSL20} and \citet{YolcuP19} used REINFORCE to train variable branching heuristic for \emph{quantified Boolean formulas} and local search algorithm WalkSAT, respectively. \citet{kurin2019improving} used DQN for SAT and \citet{Vaezipoor_Lederman_Wu_Maddison_Grosse_Seshia_Bacchus_2021} improved a SOTA \#SAT solver via \emph{Evolution Strategy}. We refer to \citep{cappart2021combinatorial, bengio2021machine} for more comprehensive record of efforts in this area.

\textbf{Unsupervised Learning for CO.}
\citet{toenshoff2021graph} proposed an unsupervised approach to solve constrained optimization problems on graphs by minimizing a problem dependent loss function. \citet{Amizadeh2019LearningTS, amizadeh2019pdp} solved SAT and CircuitSAT problems by minimizing an energy function. Lastly, inspired by probabilistic method, \citet{karalias2020erdos} trained a GNN in an unsupervised way to act as a distribution over possible solutions of a given problem, by minimizing a probabilistic penalty loss. Our method is distinct from other unsupervised techniques in that we do not minimize a problem-dependent loss function and rather learn problem representations through contrastive learning. To the best of our knowledge this is the first attempt at applying contrastive learning in the domain of combinatorial optimization.

\section{Framework Overview}

\begin{figure*}[th]
    \centering
    \begin{tikzpicture}
    \node at (0,0) {\includegraphics[width=2\columnwidth]{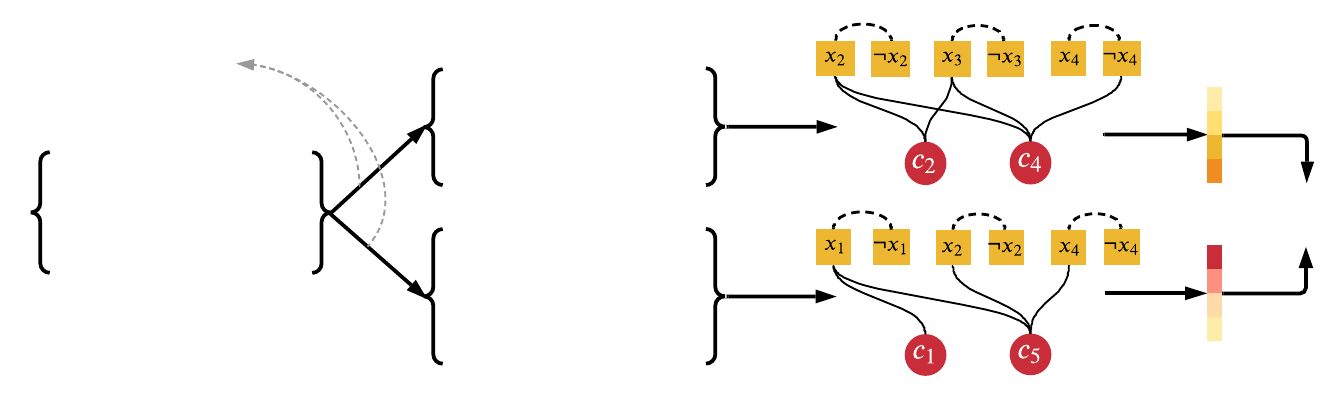}};

    \scriptsize
    \draw (-7.7, 0.4)  node[right] {$c_1:x_1$};
    \draw (-7.7, 0.4 - 0.35)  node[right] {$c_2: x_2 \lor x_3 $};
    \draw (-7.7, 0.4 - 2 * 0.35)  node[right] {$c_3: x_1 \lor \neg x_3 \lor x_4 $};
    \draw (-7.7, 0.4 - 3 * 0.35)  node[right] {$c_4: \neg x_1 \lor x_2 \lor x_3 \lor \neg x_4$};
    
    \draw (-2.8, 1.45)   node[right] {\st{$c_1: x_1$}};
    \draw (-2.8, 1.45  - 0.35)  node[right] {$c_2: x_2 \lor x_3$};
    \draw (-2.8, 1.45  - 2 * 0.35) node[right] {\st{$c_3: x_1 \lor \neg x_3 \lor x_4 $}};
    \draw (-2.8, 1.45  - 3 * 0.35)  node[right] {$c_4\!\!: $\st{$\neg x_1 \lor$} $ x_2 \lor x_3 \lor \neg x_4$};
    
    \draw (-2.8, -0.55)   node[right] {$c_1: x_1$};
    \draw (-2.8, -0.55 - 0.35)  node[right] {\st{$c_2: x_2 \lor x_3$}};
    \draw (-2.8, -0.55 - 2 * 0.35)  node[right] {\st{$c_3: x_1 \lor \neg x_3 \lor x_4 $}};
    \draw (-2.8, -0.55 - 3 * 0.35)   node[right] {\st{$c_4: \neg x_1 \lor x_2 \lor x_3 \lor \neg x_4$}};
    \draw (-2.8, -0.55 - 4 * 0.35)   node[right] {\color{our_red} $c_5: x_1 \lor x_2 \lor x_4$};

    \footnotesize
    \draw (-5.3, 2.)   node[left] {Label};
    \draw (-5.3, 1.65)  node[left] {Preserving};
     \draw (-5.3, 1.35)  node[left] {Augmentations};
    
    \normalsize
    \draw (5.8, 0.55) node {Encoder};
    \draw (5.8, -1.4) node {Encoder};
    
    \draw (2.5, 0.55) node {$G_1$};
    \draw (2.5, -1.7) node {$G_2$};
    
    \draw (1.2, 0.65)   node {$\mathcal{E}$};
    \draw (1.2, -1.45)   node {$\mathcal{E}$};
    
    \draw (6.8, 0)   node {$\mathbf{z_1}$};
    \draw (6.8, -1.95)   node {$\mathbf{z_2}$};
    
    \large
    \draw (-6.0, -1.1)   node {$\phi$};
    \draw (-1.2, -0.1)   node {$\hat{\phi}_1$};
    \draw (-1.2, -2.4)  node {$\hat{\phi}_2$};
    
    \Large
    \draw (7.8, -0.25)   node {$\mathcal{L(.)}$};

 \end{tikzpicture}
    \caption{Our contrastive learning framework for combinatorial problems. Given an instance $\phi$, a pair of augmented samples $(\hat{\phi}_1$, $\hat{\phi}_2)$ are formed using label-preserving augmentations. $(\hat{\phi}_1$, $\hat{\phi}_2)$ are then transformed to graph formats $(G_1, G_2)$. An encoder is used to extract graph representations $(\mathbf{z}_1, \mathbf{z}_2)$. Lastly, a standard contrastive loss is applied over a mini-batch of instances.} 
     \label{fig:cocl_model}
\end{figure*}
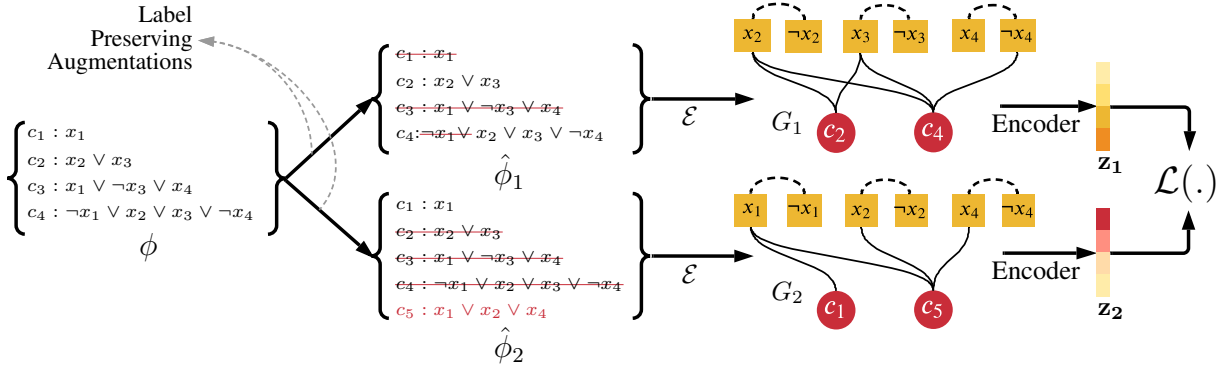

Similar to contrastive algorithms in other domains (e.g., images), we learn representations by contrasting augmented views of the same instance against negative samples. 
Our framework (Figure \ref{fig:cocl_model}) consists of four major components:
    
    \textbf{Augmentations.}
    Given a combinatorial instance $\phi$, a stochastic augmentation is applied to form a pair of positive samples, denoted by $(\hat{\phi}_1$, $\hat{\phi}_2)$. Our key insight is the tailored design of augmentations for combinatorial problems should preserve the label of the problem, e.g., satisfiability for SAT, as detailed in Section \ref{sec: lpa}.

    \textbf{Format Transformation ($\mathcal{E}$).}
    Formulas $(\hat{\phi}_1$, $\hat{\phi}_2)$ are transformed into (e.g., $\text{LIG}^{+}$) graphs $(G_1, G_2)$.

    \textbf{Encoder.}
    A neural encoder is trained to extract graph-level representations $\mathbf{z}_1, \mathbf{z}_2 \in \mathbb{R}^d$ from the augmented graphs $G_1$ and $G_2$. This encoder can be any GNN such as GCN \citep{kipf2016semi} and the encoder proposed by NeuroSAT.
    
    \textbf{Contrastive Loss.} 
    In the end, SimCLR's \citep{chen2020simple} contrastive loss $\mathcal{L}$ is applied to the graph representations $\{ \mathbf{z}_i \}_{i=1}^{2n}$, obtained from a mini-batch of $n$ instances. 
    We follow \citet{chen2020simple} and use an MLP projection head to project each $\mathbf{z}_i$ to $\mathbf{m}_i$. For a positive pair of projected representations $(\mathbf{m}_i, \mathbf{m}_j)$, the loss $\mathcal{L}_{i, j}$ is a cross-entropy loss of differentiating the positive pair from the other $2(n-1)$ negative samples
    (i.e., augmented samples of other instances): 
    \begin{align}
        \mathcal{L}_{i, j} = -\log \frac{\exp(sim(\mathbf{m}_i, \mathbf{m}_j)\mathbin{/} \tau)}{\sum_{k=1}^{2N} \mathbf{1}_{k \neq i} \exp(sim(\mathbf{m}_i, \mathbf{m}_k)\mathbin{/} \tau)},
        \label{equ: contrastive}
    \end{align}
    where $\tau$ is the temperature parameter, $\mathbf{1}$ is the indicator function, and $sim$ measures the similarity between two representations: $sim(\mathbf{m}_i, \mathbf{m}_j) := \mathbf{m}_i^T \mathbf{m}_j / \left\Vert \mathbf{m}_i \right\Vert \left\Vert \mathbf{m}_j \right\Vert$. The final loss $\mathcal{L}$ is the average of $\mathcal{L}_{i, j}$ over all positive pairs. 
    After training is completed, we only keep the encoder to extract the representation $\mathbf{z}_i$ for downstream tasks. 

\section{Augmentations for Combinatorial Problems}
\label{sec: lpa}
\subsection{Definition and Motivation}
Intuitively, contrastive learning leads to representations that are invariant across augmentations. Thus, to guarantee downstream predictive performance, augmentations should (mostly) preserve the labels of downstream tasks \citep{arora2019theoretical, tosh2021contrastive, haochen2021provable, dubois2021lossy}. 
The augmentations found in computer vision, e.g., cropping and color jittering of images, typically preserve the labels of classification tasks.
This is not the case for most previous graph contrastive frameworks \citep{you2020graph, hassani2020contrastive}, where simple graph augmentations like node dropping and link perturbation are used without consideration of the downstream task. For combinatorial problems, these \emph{label-agnostic augmentations} (\laas) are likely to produce many false positive pairs. For example, in the SR dataset in NeuroSAT, each SAT can be turned into UNSAT by flipping one literal in one clause. 

Thus, we aim to design \emph{label-preserving augmentations} (\lpas), which are transformations that preserve the instance label. Formally, given a problem family $\Phi$ and a labelling function $f$, an augmentation distribution $A$ is an \lpa iff:
\begin{align*}f(\hat{\phi}) = f(\phi), \quad \forall \hat{\phi} \in \supp\left(A(\cdot | \phi)\right), \forall \phi \in \Phi,\end{align*}

where $\supp$ denotes the support of a distribution.

Importantly, \lpas are well-studied for many combinatorial problems and are much cheaper to obtain than labels. For SAT prediction, common preprocessing techniques from SAT solvers such as variable elimination can be used as \lpas, as discussed in Section \ref{subsec: augmentation_sat}. There are also \lpas for other combinatorial problems, such as, adding cuts \cite{achterberg2020presolve} for MILPs and deleting dominant vertices \citep{akiba2016branch} for Minimum Vertex Cover.

The type of \lpas is also crucial. Intuitively, \lpas that make more significant changes to an instance create harder positives from which better representations can be learned. Indeed, \lpas that lead to larger augmentation support $\supp\left(A(\cdot | \phi)\right)$ will split $\Phi$ into coarser equivalences classes, and thus (roughly speaking) a classifier can be learned with fewer labelled instances.

\subsection{Label-preserving Augmentations for SAT}
\label{subsec: augmentation_sat}

The \lpas for {SAT} preserve satisfiability of any Boolean formula. In other words, 
a \textit{(un)}satisfiable instance remains \textit{(un)}satisfiable after the applications of \lpas. We review some common LPAs for SAT below, with examples and time complexity results provided in Appendix \ref{appendix: example_lpas}.

\textbf{Unit Propagation (UP).} A clause is a \emph{unit clause} if it contains only one literal. If an instance $\phi$ contains a unit clause $\ell$, we can 1) remove all clauses in $\phi$ containing the literal $\ell$ and 2) delete $\neg \ell$ from all other clauses.

\textbf{Add Unit Literal (AU).} The inverse of UP: 1) construct a unit clause from a new literal $\ell$, 2) add its negation $\neg \ell$ to some other clauses and 3) create new clauses containing $\ell$. 

\textbf{Pure Literal Elimination (PL).} A variable $v$ is called pure if it occurs with only one polarity in $\phi$. We can delete all clauses in $\phi$ containing $v$.  
    
\textbf{Subsumed Clause Elimination (SC).} If a clause $c_1$ is a subset of $c_2$, i.e., all literals in $c_1$ are also in $c_2$, then deleting $c_2$ does not change satisfiability of $\phi$.

\textbf{Clause Resolution (CR).} Resolution produces a new clause implied by two clauses containing complementary literals:
    \begin{equation}
        \frac{\ell \lor a_1 \lor  \cdots \lor a_n, \neg \ell \lor b_1 \lor \cdots b_m   }{a_1 \lor  \cdots \lor a_n \lor b_1 \lor \cdots b_m  }
        \label{equ: resolution}
    \end{equation}
The new clause $c$ is called the \textit{resolvent} of $c_1$, $c_2$: $c = c_1 \otimes c_2$. Adding $c$ to $\phi$ does not change satisfiability.

\textbf{Variable Elimination (VE).} 
Let $S_\ell$ be a set of clauses containing the literal $\ell$, and $S_{\neg \ell}$ be a set of clauses containing its negation $\neg \ell$. Then a new set $S$ is obtained by pairwise resolving on clauses of $S_\ell$ and $S_{\neg \ell}$: $S = \{c_1 \otimes c_2 | c_1 \in S_\ell, c_2 \in S_{\neg \ell}\}$. Replacing $S_\ell \cup S_{\neg \ell}$ with $S$ does not change satisfiability \cite{een2005effective}.

Note that some augmentations, such as VE, have the worst-case exponential complexity if run until convergence. However, in our paper, we only eliminate a small and fixed number of variables. Therefore, all LPAs listed here are cheap and have polynomial-time complexity.

\section{Empirical Study of Augmentations for SAT}
\label{sec: augmentation}
\begin{figure*}[htbp]
    \centering

    \begin{tikzpicture}
    \node at (0,0) {\includegraphics[width=.85\textwidth]{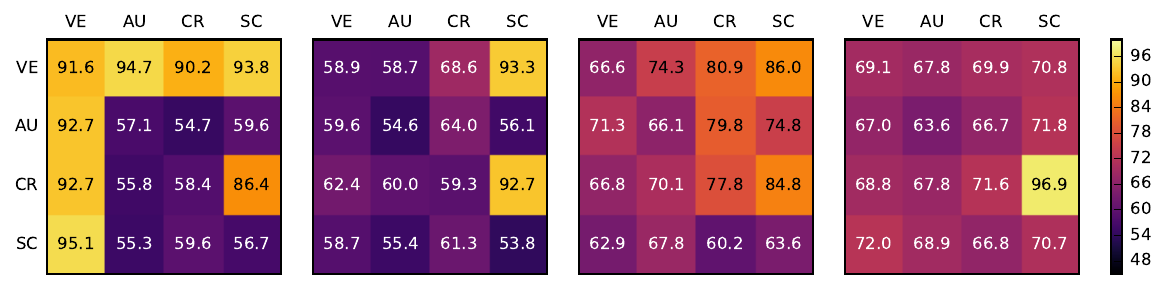}};
    \node at (0,-3.2) {\includegraphics[width=.85\textwidth]{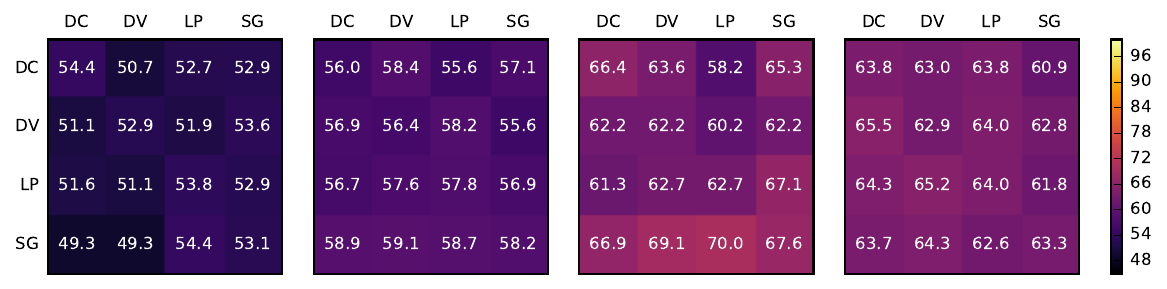}};
    
    \footnotesize
    \draw (-5.2, -5.2)  node {SR(10)};
    \draw (-1.8, -5.2)  node {PR(10)};
    \draw (1.4, -5.2)  node {DP(20)};
    \draw (4.8, -5.2)  node {PS(20)};
    
    \node[rotate=90] at (-7.6, -0.2) {LPA};
    \node[rotate=90] at (-7.6, -3.5) {LAA};
    \end{tikzpicture}
    
    \caption{Contrastive learning with \lpas learns much better representations for SAT predictions than with \laas. 
    We evaluated the linear classification performance of SSL models trained with different single augmentations (diagonal) or paired combinations (off-diagonal). 
    The four heatmaps on the first row show results using \lpas, and the ones on the second row using \laas.
    Each column denotes different datasets.
    For off-diagonal entries on each heatmap, the row corresponds to the first augmentation applied. 
    }
    \label{fig: heatmap}
\end{figure*}
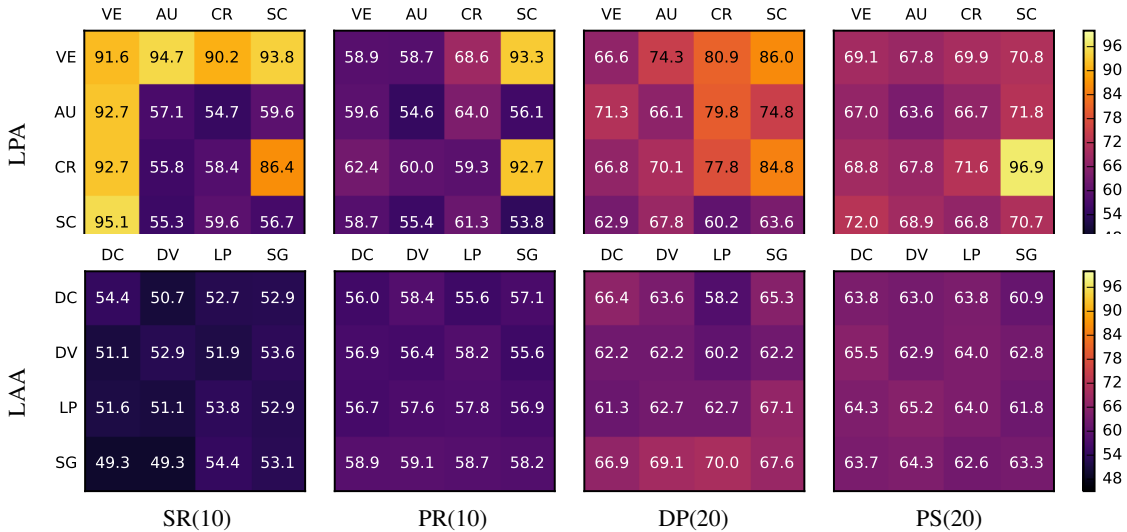

Our first experiments study the role of augmentations in contrastive learning for SAT prediction. We followed the standard linear evaluation protocol to evaluate representations \citep{chen2020simple}, i.e., we report test accuracy of a linear classifier trained on top of frozen representations. 

\textbf{Architecture.} We primarily used the encoder of NeuroSAT  \citep{selsam2018learning} as the GNN architecture. We also re-ran a small number of experiments with another type of GNN in Appendix \ref{appendix: other_gnn}. The dimension of literal representations was chosen to be $128$. We discarded the aggregator of NeuroSAT and obtained the graph-level representations by average-pooling over all literal representations. All design details of the encoder followed the original NeuroSAT paper unless otherwise specified. 

\textbf{Experimental Setting.}   
We used the contrastive loss in Equation \ref{equ: contrastive} with the temperature $0.5$. For the projection head, we used a 2-layer MLP, with the dimension of hidden and output layer being $64$. Appendix \ref{appendix: vicreg} also shows some experiments with other contrastive objectives. We used Adam optimizer with learning rate $2 \times 10^{-4}$ and weight decay $10^{-5}$. The batch size was $128$ and the maximum training epoch was $5000$. The generator produced a set of new unlabelled instances for each batch. We used sklearn's \citep{scikit-learn} logistic regression model for linear evaluation. We generated 100 separate labelled instances to train our linear evaluators, and another $500$ as the validation set to pick the hyperparameters (ranging from $10^{-3}$ to $10^3$) of $L_2$ regularization. The test set consisted of $10^4$ instances. 

\textbf{Datasets.} We experimented using four generators: SR \citep{selsam2018learning}, Power Random 3SAT (PR) \citep{ansotegui2009towards}, Double Power (DP) and Popularity Similarity (PS) \citep{giraldez2017locality}. SR and PR are the synthetic generators. DP and PS are pseudo-industrial generators producing instances that mimic real-world problems. We generated instances of 10 variables for SR and PR, and 20 for DP and PS. The number inside the parenthesis denotes the number of variables per instance. We also tweaked the parameters of the generators so that they produce roughly balanced SAT and UNSAT. More details are in Appendix \ref{appendix: dataset}.

\textbf{Augmentations.} We used four of the LPAs from Section \ref{subsec: augmentation_sat}, namely: \au, \satsc, \satcr  and \ve. The augmentations UP and PL were not studied because most of our instances originally do not contain unit clauses or pure literals. 

For our baseline we adopted four \laas from GraphCL \citep{you2020graph} with some adjustments. The issue with the original augmentations was that they operate directly on the graph. However, NeuroSAT requires the input to be in the $\text{LIG}^{+}$ format, and blindly applying these augmentations might break that structure. Thus, we adapted GraphCL augmentations to maintain the $\text{LIG}^{+}$ structure, resulting in the following augmentations: \textit{drop clauses} (\dc), \textit{drop variables} (\dv), \textit{link perturbation} (\lp) and \textit{subgraph} (\sg). \dc and \dv correspond to \textit{node dropping}. LP corresponds to \textit{edge perturbation}, where we randomly add or remove links between literals and clauses on a SAT instance. Lastly, SG is similar to the one in GraphCL where we do a random walk on an instance and keep the resulting subgraph.

All $8$ augmentations except \satsc are parameterized by $p$ which controls the intensity of perturbations. For \satsc, we eliminate all subsumed clauses. In this section, we chose the $p$ that achieved the best linear evaluation performance 
when the corresponding augmentation was applied alone.  The results for tuning $p$ are shown in Appendix \ref{appendix: hyperparameter}.

\subsection{Label-Preserving Augmentations are Necessary}

\begin{table}[tb]
\caption{Adding SAT-preserving clauses (Resolution) leads to much higher accuracy than adding random ones (Random). The number of clauses added is controlled to be the same for both.}
\label{tab: random_resolution}
\small
\centering
\begin{tabular}{l|cccc}
\toprule
\diagbox{Type}{Dataset} & SR & PR & DP & PS \\
\midrule
Random & 55.6 & 52.8 & 64.7 & 68.9 \\
Resolution (ours) & \textbf{86.4} & \textbf{92.6} & \textbf{84.8} & \textbf{96.9} \\
\bottomrule
\end{tabular}
\end{table}

\textbf{Do \lpas \ learn better representations than \laas?} 
To investigate the effect of using different augmentations, we evaluated the linear classification performance of SSL models trained with different single augmentations or paired combinations.
As Figure \ref{fig: heatmap} shows, 
the accuracy ($\%$) for the best LPA pair is: $95.1$ for SR, $93.3$ for PR, $86.0$ for DP and $96.9 $ for PS, while the corresponding number ($\%$) for \laas is much lower: $54.4,  59.1, 70.0$ and $65.5$. These results show that SSL models trained with \lpas learn significantly better representations than those with \laas. 

\textbf{Do our gains simply come from adding clauses?} The four LAAs from GraphCL do not add new clauses to the original instance. To investigate whether this explains the performance gap, we compared \satcr against adding the same number of randomly generated clauses (\textit{adding random clauses}). We eliminated subsumed clauses (SC) after both augmentations. As shown in Table \ref{tab: random_resolution}, adding SAT-preserving resolvents achieved at least $20 \%$ higher accuracy than adding randomly generated clauses.

\textbf{Does combining \lpas \ and \laas \ hurt performance?} In Table \ref{tab: nspp_after_spp}, we trained SSL models with \ve \ followed by $7$ different augmentations on SR(10). We compared the accuracy change with the model trained with \ve \ only. From Table \ref{tab: nspp_after_spp}, adding LAAs resulted in $30 \%$ drop in accuracy, while LPAs decreased the accuracy by at most $3 \%$.

In summary, the performance gap between LPAs and LAAs meets our intuition.  LAAs do not guarantee preserving satisfiability, which could result in false positive pairs that hurt the SAT prediction performance. 

\begin{table}[tb]
\caption{Adding \laas to \lpas significantly hurts the performance. Cells represent the difference of linear evaluation accuracy between training with VE followed by another aug and with VE alone on SR(10). Yellow indicates improved accuracy.}
\label{tab: nspp_after_spp}
\centering
\small
\begin{tabular}{ccccccc}
\toprule
  \multicolumn{3}{c}{\lpas} & \multicolumn{4}{c}{\laas} \\
  \cmidrule(lr){1-3} \cmidrule(lr){4-7}
  \au & \satcr & \satsc & \dc & \dv & \lp & \sg \\ 
        \midrule
 \ccar{-1.5} & \ccay{0.7} & \ccay{1.2} & \textbf{\ccar{-34.4}} & \textbf{\ccar{-32.4}} & \textbf{\ccar{-33.7}} & \textbf{\ccar{-35.9}} \\
\bottomrule
\end{tabular}
\end{table}

\subsection{Type, Order, and Strengths of LPAs are Crucial}

Previous studies of contrastive learning for image \citep{chen2020simple} and graph \citep{you2020graph} have shown that the quality of learned representations relies heavily on finding the right type and composition of augmentations. We observe a similar pattern in  SAT. In Figure \ref{fig: heatmap}, the accuracy gap between the best LPA combination and the worst is between $20 - 40 \%$ for all datasets.

The conjecture in SimCLR \citep{chen2020simple} is that stronger augmentations lead to better representations. Intuitively, weaker augmentations often create very correlated positive examples, providing shortcuts for neural networks to cheat in the contrastive task without learning meaningful representations. 
Based on this conjecture, we study what type, order, and strengths of LPAs induce harder positive pairs and thus better representation quality.

\textbf{Resolution-based augmentations are the most powerful.} In Figure \ref{fig: heatmap}, the best pair for each dataset includes either \satcr \ or \ve . Both of them are based on the resolution rule in Equation \ref{equ: resolution}. 
Resolution is a powerful inference rule in propositional logic. In fact, we can build a sound and complete propositional theorem prover with only resolutions \citep{genesereth2013introduction}. The Davis–Putnam algorithm \citep{davis1960computing}, the basis of practical SAT solvers, iteratively applies resolution until reaching satisfiability certificates. Resolution-based augmentations in contrastive learning may help the neural work learn the essence of resolution, which leads to better satisfiability prediction.

\textbf{Composing different augmentations is beneficial across datasets.} The highest accuracy in Figure \ref{fig: heatmap} always comes from off-diagonal entries, i.e., composition of different augmentations. In PR, every singular augmentation failed to obtain accuracy higher than $60 \%$ by itself. 
While singular augmentations achieved decent accuracy on SR and DP, combinations further improved the accuracy. Similar to image and graph, composing different augmentations resulted in harder positives and better representations. 

\textbf{Eliminating subsumed clauses after adding resolvents is particularly helpful.} \satcr followed by \satsc \ consistently produced high-quality representations across all datasets. Using \satcr \ alone tended to perform substantially worse. The accuracy drop without \satsc is: $28.1 \%$ for SR, $33.4 \%$ for PR, $7.0 \%$ for DP, $25.3 \%$ for PS. Without \satsc, all original clauses are preserved by the augmentation, leading to trivial positive pairs. In other words, without \satsc the GNN may have solved the contrastive task by finding a common subgraph, not by learning anything meaningful about resolution rules.

Interestingly, swapping the order of this pair also decreased accuracy by $20-30 \%$ for all datasets. We conjecture that this may be an artefact of our specific generators. The percent of subsumed clauses in the original instances (over $1000$ samples) was $43, 0, 76, 69$ for SR, PR, DP and PS. Applying \satsc first had no effect on PR due to no subsumed clauses present. For the other three, \satsc may have eliminated too many clauses, which hurt the diversity of resolvents created in the next phase.

\begin{figure}[bt]
    \centering
    \begin{tikzpicture}
    \node at (0,0) {\includegraphics[width =.3\textwidth]{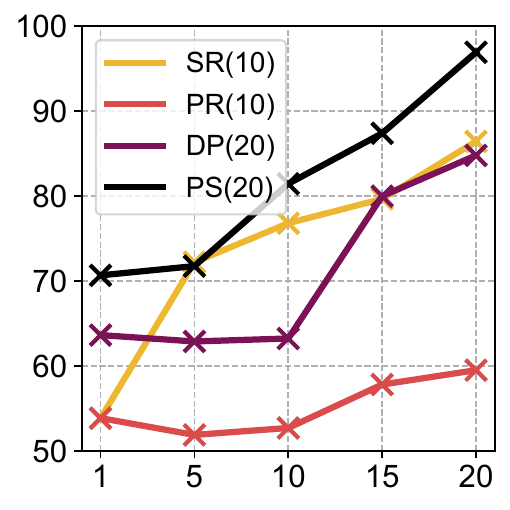}};
    
    \footnotesize
    \node[bag, rotate=90] at (-2.8, 0) {Linear Evaluation \\ Accuracy (\%)};
    \draw (0.4, -2.8)  node {Rate of Clause Resolution (\%)};
    \end{tikzpicture}
    \caption{Adding more resolvents improves the performance across all datasets. Each point represents linear evaluation accuracy of an SSL model trained with \satcr \ of the specified rate (i.e., \# of added resolvents divided by \# of original clauses) followed by \satsc.}
    \label{fig: more_clause}
\end{figure}

\textbf{More resolvents lead to better representations.} Figure \ref{fig: more_clause} studies the effect of the number of resolvents added in \satcr. In general, adding more resolvents improved the linear evaluation performance across all datasets. More resolvents not only add more new clauses, but also may help \satsc \ delete more original clauses, which create harder positives. 

\subsection{Do Our Augmentations Reveal the True Label?}

As mentioned before, \satcr and \ve are powerful enough to build a sound and complete solver. Because our instances are relatively small, it is important to ask if our augmentations ``accidentally'' reveal the true label(SAT/UNSAT) to the model. We do not believe this is the case for the following reasons.

Our augmentations (polytime complexity) cannot in general determine whether a formula is satisfiable (exponential in the worst case). When applied to our datasets, \ve  only eliminated between 1-4 variables, and \satcr added between 10-20\% of the total number of original clauses. This amount of work alone is not enough to solve SAT.

To assess whether our augmentations are effectively solving our \emph{specific} instances, we measured the number of decision steps of CryptoMiniSat solvers \cite{soos2009extending} on different datasets before and after our \lpas. If our LPAs were close to solving the instance, the decision steps should have decreased dramatically after LPAs were applied. However, Table \ref{table: decision_step} in Appendix \ref{appendix: decision} shows the decision steps remain relatively stable after augmentations.

\begin{figure*}[tb]
    \centering
    
    \begin{tikzpicture}
    \node at (0,0) {\includegraphics[scale=0.41]{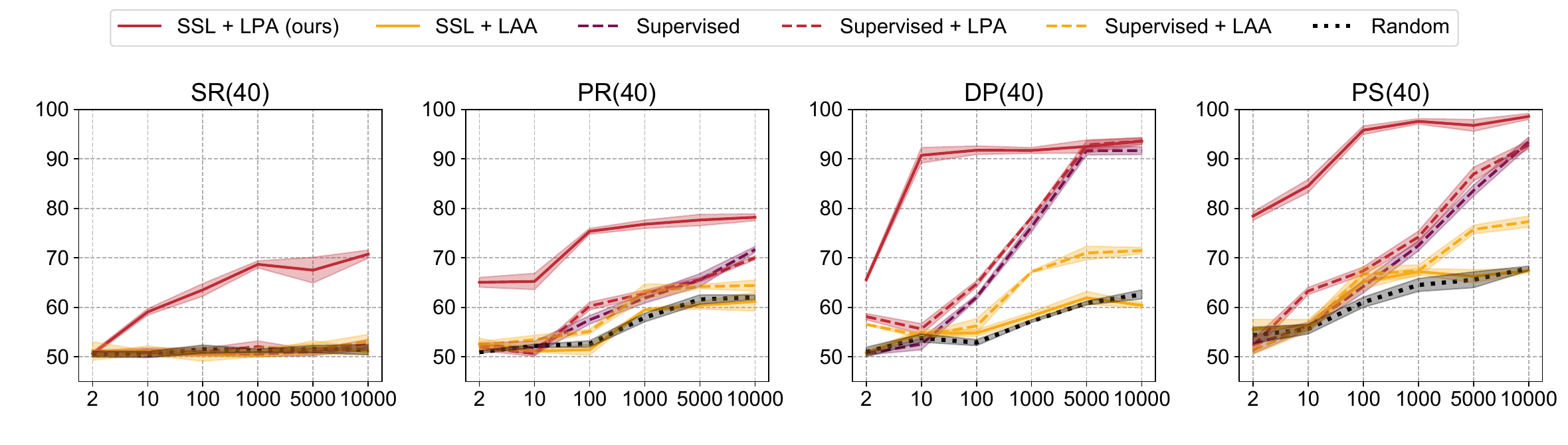}};
    \node at (0,-4) {\includegraphics[scale=0.41]{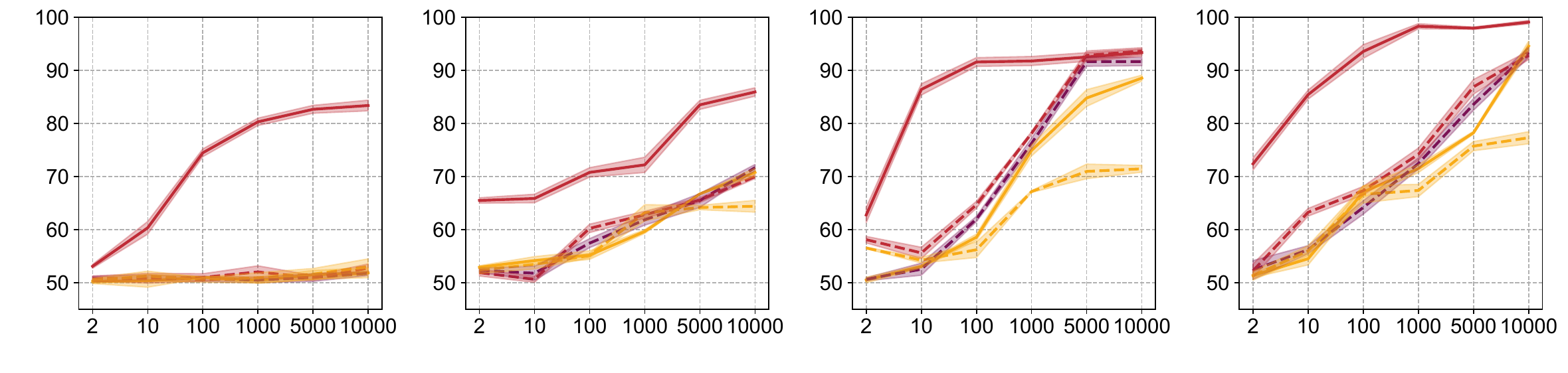}};
    \node[bag, rotate=90] at (-8.4, -0.2) {Linear Evaluation};
    \node[bag, rotate=90] at (-8.4, -3.5) {Fine-Tuning};
    \scriptsize
    \node[bag, rotate=90] at (-8, -0.2) {Accuracy (\%)};
    \node[bag, rotate=90] at (-8, -3.5) {Accuracy (\%)};
    
    \draw (-5.75, -5.8)  node {No. of Training Labels};
    \draw (-1.75, -5.8)  node {No. of Training Labels};
    \draw (2.5, -5.8)  node {No. of Training Labels};
    \draw (6.5, -5.8)  node {No. of Training Labels};
    
    \end{tikzpicture}
    \caption{Our method (SSL + LPA) achieves significantly higher accuracy after linear evaluation and fine-tuning than all baselines in low-label regime and is comparable to supervised models that have been given more labels. We vary the number of training labelled instances from $2$ to $10^4$ and report the average accuracy and standard error over 3 trials for all methods. }
    \label{fig: main_result}
\end{figure*}
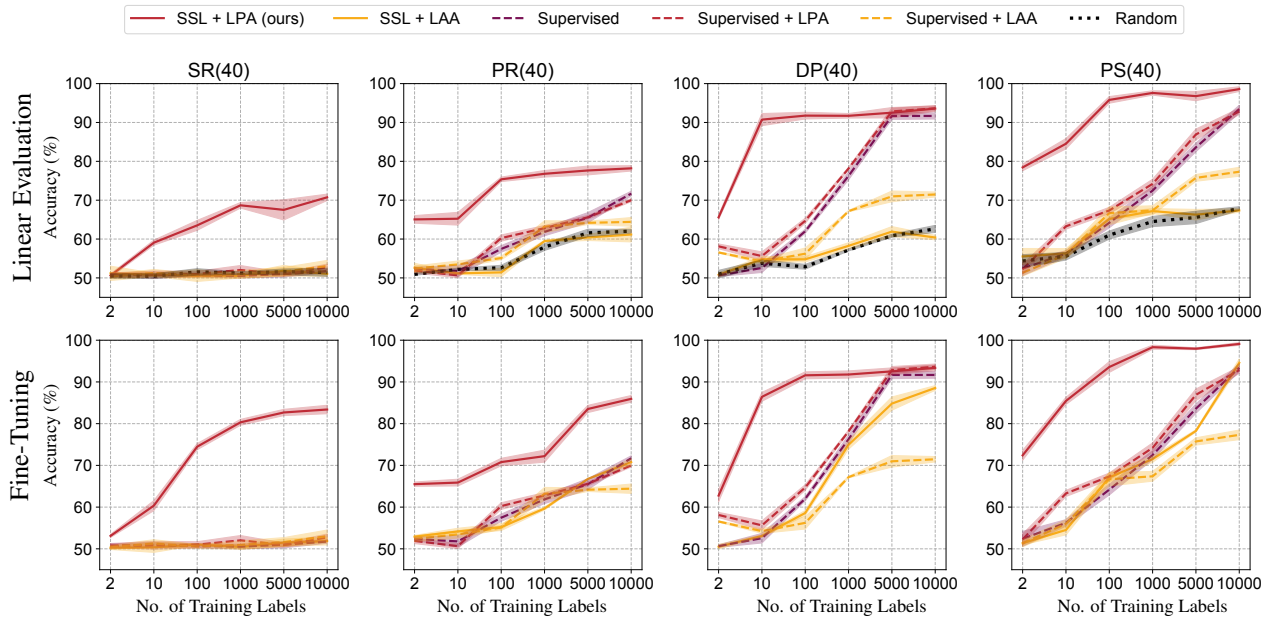

\section{Comparison with Other Methods}
\label{sec:experiment}

The second set of experiments compared our proposed framework with other baselines in the setups of linear evaluation, fine-tuning and few-shot transfer learning. In addition to our model (SSL with LPA), we studied 4 baselines: SSL with LAA, supervised models trained without augmentations, supervised models trained with LPA or LAA. The details of training SSL models followed Section \ref{sec: augmentation}. For supervised models, we trained the encoder and aggregator of NeuroSAT end-to-end. The hyperparameters for supervised models were the same as SSL in Section \ref{sec: augmentation}, except the learning rate was chosen to be $2 \times 10^{-5}$, following \citet{selsam2018learning}. For each dataset, we chose the best augmentation combination for LPA and LAA according to Figure \ref{fig: heatmap}. The degree parameter associated with each augmentation was tuned separately for SSL and supervised models. We used $200$ instances as validation sets for early stopping of all methods.

Unless otherwise specified, all datasets used in this section have 40 variables per instance; in the Appendix \ref{appendix: eval_small} we report results for the same experiments on smaller instances. For SR datasets, following NeuroSAT, we trained on SR(U(10, 40)) and tested on SR(40). Training procedures for all models were the same as Section \ref{subsec: experiment_linear}, except the batch size was $80$ for SR(40) to avoid memory issues. When fine-tuning NeuroSAT, we used different learning rates for the encoder and aggregator, which were separately tuned for different models on each dataset.

\subsection{Linear Evaluation}
\label{subsec: experiment_linear}
We first evaluated the linear evaluation accuracy of all methods following the procedure in Section \ref{sec: augmentation}. 
We varied the number of labelled instances from $2$ to $10^4$.

As shown in the first row of Figure \ref{fig: main_result}, our method achieved substantially higher accuracy than others across all datasets in the low-label regime. For example, with $10$ training labelled instances, the improvement ($\%$) of ours compared from the second best method was: $9.42$ for SR, $14.07$ for PR, $30.81$ for DP, $22.18$ for PS. Our model's accuracy was also on par with fully-supervised models that had access to significantly more labels: for all datasets, the accuracy of our models trained with $100$ labels matched or exceeded the accuracy of our best supervised baselines trained with $10^4$ labels, a $100\times$ reduction in the number of labels needed.

On the other hand, SSL models trained with LAAs were not much better than random-initialized ones under linear evaluation. We also found that using LAAs for supervised models even hurt performance, possibly because LAAs add label noise. 
For example, with $10000$ labels on DP(40), adding LAAs gave $20.21 \%$ lower accuracy than supervised without augmentations.

\subsection{Fine-tuning}

ƒWe evaluated the fine-tuning accuracy of different SSL models compared to the supervised baselines. Specifically, we took the pre-trained SSL models and optimized them end-to-end on labelled data for a few epochs.

As shown in the second row of Figure \ref{fig: main_result}, the fine-tuning results resemble those of linear evaluation. In the low-label regime, our model (SSL + LPA) dominated across all datasets. In contrast, SSL with LAAs did not improve much from supervised models that were randomly initialized. As with linear evaluation, we observed a reduction in sample complexity of at least $100\times$. 

\subsection{Few-shot Transfer Learning}

\begin{figure}[bt]
    \centering
    \begin{tikzpicture}
    \node at (0,0) {\includegraphics[width =.3\textwidth]{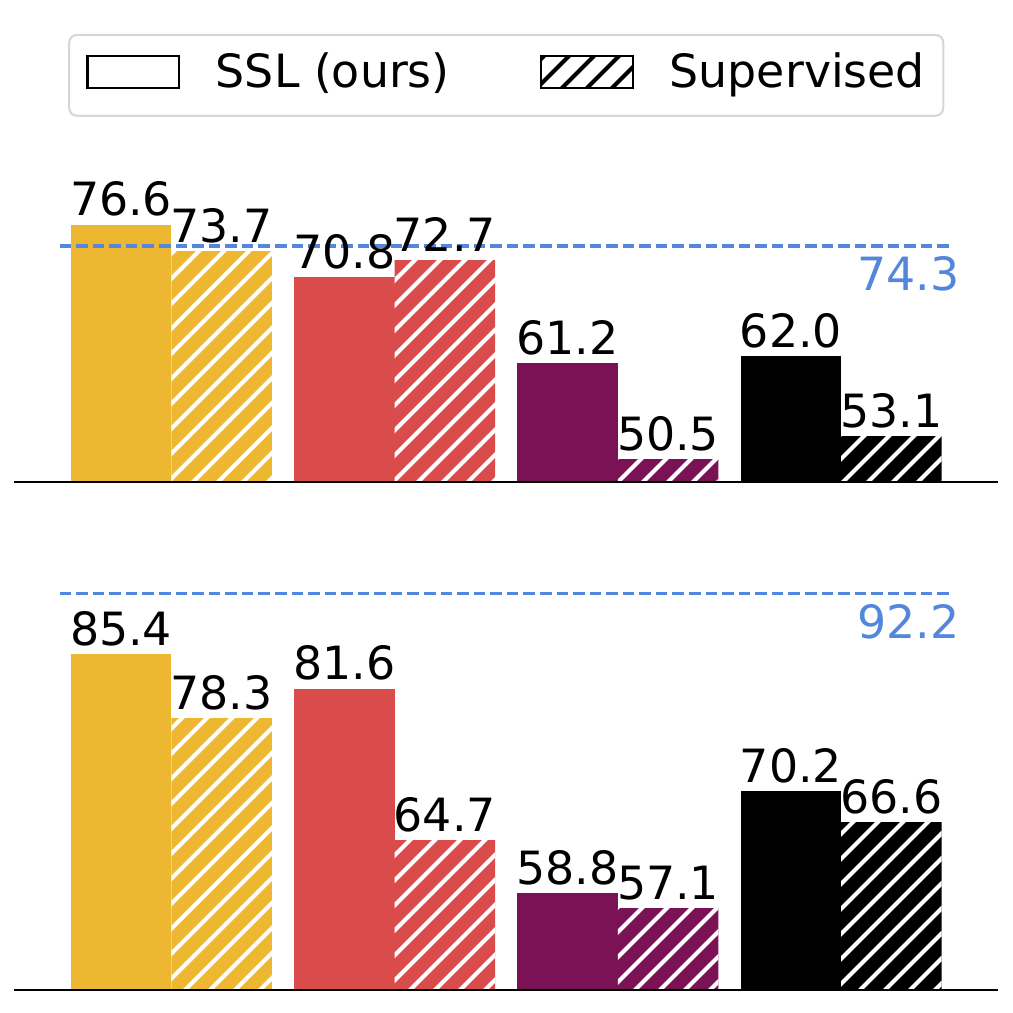}};
    
    \node[rotate=90] at (-3.2, -0.4) {Target Dataset};
    \draw (0, -3.2)  node {Source Dataset};
    
    \footnotesize
    \node[rotate=90] at (-2.6, 0.8) {UR(40)};
    \node[rotate=90] at (-2.6, -1.5) {CA(40)};
    
    \draw (-1.7, -2.7)  node {SR(10)};
    \draw (-1.7 + 1.1, -2.7)  node {PR(10)};
    \draw (-1.7 + 1.1 * 2 + 0.1, -2.7)  node {DP(20)};
    \draw (-1.7 + 1.1 * 3 + 0.1, -2.7)  node {PS(20)};
    
    \end{tikzpicture}
    \caption{The representations of our model (solid bars) transfer better from source to unseen target dataset than those of supervised (hatched bars) trained on the same source. 10-shot transfer accuracy (\%) was evaluated for each setup. The blue line is the accuracy of fully-supervised models trained on each target dataset with $10k$ labels.}
    \label{fig: transfer_supervised}
\end{figure}

We investigated how well the SSL models trained in Section \ref{subsec: experiment_linear} perform in 10-shot transfer to unseen and larger datasets, Uniform Random 3SAT (UR) and Community Attachment (CA) \citep{giraldez2015modularity}. UR is a synthetic generator, and CA is pseudo-industrial. We trained a logistic regression classifier on the fixed representations with $10$ SAT and $10$ UNSAT instances from the target dataset. 
We compared transferability of representations from SSL with those from supervised representations trained on the same source dataset. In particular, the supervised baseline was trained using $100,000$ labelled source instances with the same augmentations as SSL. Only the encoder of the supervised models was used for extracting representations on target datasets. 

\textbf{Do our representations transfer better than supervised?} As shown in Figure \ref{fig: transfer_supervised}, we found that our SSL models generally transferred better than the supervised baseline. This improvement also tended to be larger when the source and target domain were more distinct, such as from pseudo-industrial instances (DP, PS) to random 3 SAT (UR), and from random 3SAT (PR) to pseudo-industrial instances (CA). This meets our intuition, because SSL does not leverage labels in the source dataset, and thus reduces overfitting on source labels \cite{yang2020transfer}.

\textbf{Does transferability vary with the source dataset?} We also found that transferability of representations depended heavily on the source datasets. Generally speaking, models trained on pseudo-industrial generators, DP and PS, transferred worse than the synthetic generators, SR and PR. We conjecture that small industrial instances are not challenging enough for models to learn generalizable representation of unseen problem domains. 
In Appendix \ref{appendix: transfer}, we also performed an experiment to show how the model trained on SR, PR, DP and PS transfer to each other, which supports similar conclusions: models trained on SR and PR transferred better to DP and PS than vice verse.

\section{Conclusions and Outlook}

We studied the effect of data augmentations on contrastive learning for the Boolean satisfiability problem. We designed label-preserving augmentations using well-studied transformations, e.g., clause resolution or variable elimination, and confirmed the hypothesis that data augmentations should be label-preserving to help in downstream prediction. The design of our augmentations was critical; we found that resolution-based augmentations, which produced more distinct augmentations, were necessary for strong results. Our contrastive method was able to learn strong SAT predictors (at least as strong as our best supervised baselines) with $100\times$ fewer labelled training instances. 

Although our results are restricted to Boolean satisfiability, they hold lessons for solver design more broadly. For example, the convex hull of MILP feasible sets is invariant to cuts, which suggests that contrastive pre-training could be used to improve Neural Diving \citep{nair2020solving}. In general, our results strongly suggest that studying and exploiting invariances can dramatically improve the sample complexity of heuristics learned via imitation learning.

The study of combinatorial problems is fruitful for the broader machine learning community, because these problems are non-trivial and so much is known about their invariances. In particular, experiments can be designed that exactly satisfy the assumptions of invariant learning theory, making combinatorial problems fantastic test beds for the emerging field of contrastive and self-supervised learning.

\section*{Acknowledgments}
We thank Roger Grosse and Guiliang Liu for valuable discussions and insights.
Resources used in preparing this research were provided, in part, by the Province of Ontario, the Government of Canada through CIFAR, and companies sponsoring the Vector Institute. We acknowledge the support of the Natural Sciences and Engineering Research Council of Canada (NSERC), RGPIN-2021-03445.

\bibliographystyle{icml2022}
\bibliography{refs}

\newpage
\appendix
\onecolumn 
\section{\lpas for SAT}
\label{appendix: example_lpas}


\begin{table}[h]
\centering
\caption{ Examples of \lpas \ for SAT. UP: remove $c_1, c_3$ and $\neg x$ from $c_4$. AU: add unit literal $\neg x_5$, add $x_5$ to $c_1$ and create a new random clauses containing $\neg x_5$  . SC: remove $c_4$ because $c_2 \subset c_4$. CR: add $c_2 \otimes c_3$. VE: eliminate $x_3$ by adding $c_2 \otimes c_3$ and $c_4 \otimes c_3$. }
\label{tab: sat_aug}
\vskip 0.15in
\begin{tabular}{|l|l|}
\hline 
 Original    &  UP  \\
     \hline 
 $c_1: x_1$    & \st{$x_1$}  \\ 
 $c_2: x_2 \lor x_3 $    &  $x_2 \lor x_3 $     \\ 
 $c_3: x_1 \lor \neg x_3 \lor x_4 $ & \st{$x_1 \lor \neg x_3 \lor x_4 $} \\ 
 $c_4: \neg x_1 \lor x_2 \lor x_3 \lor \neg x_4$ & \st{$\neg x_1 \lor$}$ x_2 \lor x_3 \lor \neg x_4$ \\
\hline
AU & SC \\
\hline 
$\color{our_red} \neg x_5$ & \\
$\color{our_red} x_5 \lor$ $x_1$ & $x_1$ \\
$x_2 \lor x_3 $ & $x_2 \lor x_3 $ \\
$x_1 \lor \neg x_3 \lor x_4 $ & $x_1 \lor \neg x_3 \lor x_4 $ \\
$\neg x_1 \lor x_2 \lor x_3 \lor \neg x_4$ & \st{$\neg x_1 \lor x_2 \lor x_3 \lor \neg x_4$} \\
$\color{our_red} \neg x_5 \lor x_1 \lor \neg x_2 \lor x_3$& \\
\hline 
CR & VE \\
\hline 
$x_1$ & $x_1$ \\
$x_2 \lor x_3 $  &   \st{$x_2 \lor x_3 $} \\ 
$x_1 \lor \neg x_3 \lor x_4 $ &  \st{$x_1 \lor \neg x_3 \lor x_4 $} \\
$\neg x_1 \lor x_2 \lor x_3 \lor \neg x_4$ &  \st{$\neg x_1 \lor x_2 \lor x_3 \lor \neg x_4$} \\
$\color{our_red} x_1 \lor x_2 \lor x_4 $ &  $\color{our_red} x_1 \lor x_2 \lor x_4 $ \\
\hline 
\end{tabular}
\end{table}



\section{NeuroSAT}\label{appendix: neurosat}

\paragraph{Encoder} We use $L^n$ and $C^n$ to denote the embeddings for literals and clauses at the message passing round $n$. In addition, we have hidden states for literals and clauses, denoted by $L_h^n, C_h^n$. Let $M$ be the bipartite adjacency matrix of the $\text{LIG}^{+}$. NeuroSAT encoder is parameterized by two MLPs ($L_{msg}, C_{msg}$) and two layer-norm LSTMs ($L_u, C_u$), which are shared for all rounds. Then for round $n$, embeddings are updated as:
\begin{align*}
    (C^{n+1}, C_h^{n+1}) & \leftarrow C_u([C_h^n, M^TL_{msg}(L^n)])\\
    (L^{n+1}, L_h^{n+1}) & \leftarrow L_u([L_h^n, FLIP(L^n), MC_{msg}(C^{n+1})]), 
\end{align*}
where $FLIP$ swaps the literal and its negation in the embedding. 

In this work, we always set the number of message passing rounds to $26$. We use $L^{26}$ as the final literal embedding. The graph-level representation can be computed as an average of the embeddings for all literals.

\paragraph{Aggregator} Given the embeddings $L^{N}$ from the encoder, we project each literal representation to a single scalar using an MLP $L_{vote}$: $V \leftarrow L_{vote} (L^N)$. Then we minimize the cross-entropy loss between the true label and $sigmoid(mean(L_{vote}))$

\section{Augmentation Rates for \laas ans \lpas}
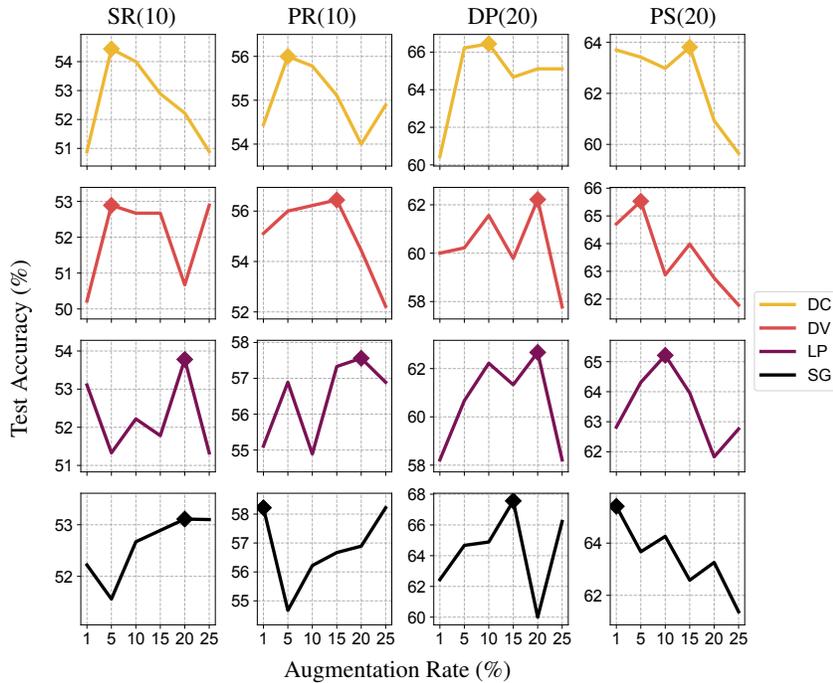
\begin{figure}[h]
    \centering
    
    \begin{tikzpicture}
    \node at (0.65,0.1) {\includegraphics[width =.62\textwidth]{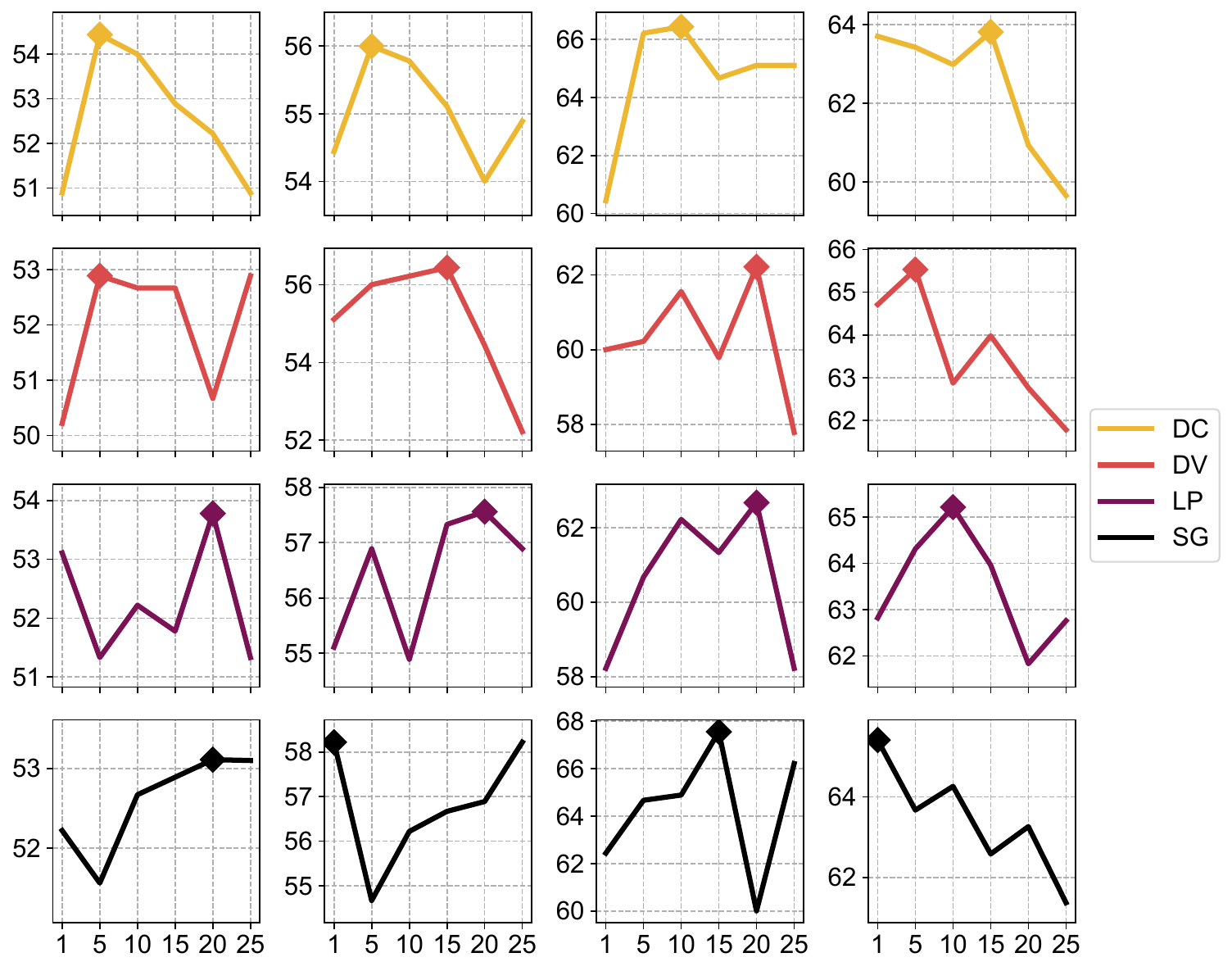}};
    
    \footnotesize
    \draw (-3.4, 4.4)  node {SR(10)};
    \draw (-3.4 + 2.4, 4.4)  node {PR(10)};
    \draw (-3.4 + 2 * 2.4, 4.4)  node {DP(20)};
    \draw (-3.4 + 3 * 2.4, 4.4)  node {PS(20)};
    \draw (0, -4.3)  node {Augmentation Rate (\%)};
    
    \node[rotate=90] at (-5, 0) {Test Accuracy (\%)};
    
    \end{tikzpicture}
    \caption{The effect of rate of LAAs on the accuracy across datasets. The diamond markers indicate the maximum achieved accuracy which was the rate that was subsequently chosen to produce the heatmap of Figure \ref{fig: heatmap}.}
    \label{fig: gcl_hyper}
\end{figure}

\begin{figure}[h]
    \centering
    
    \begin{tikzpicture}
    \node at (0.65,0.1) {\includegraphics[width =.62\textwidth]{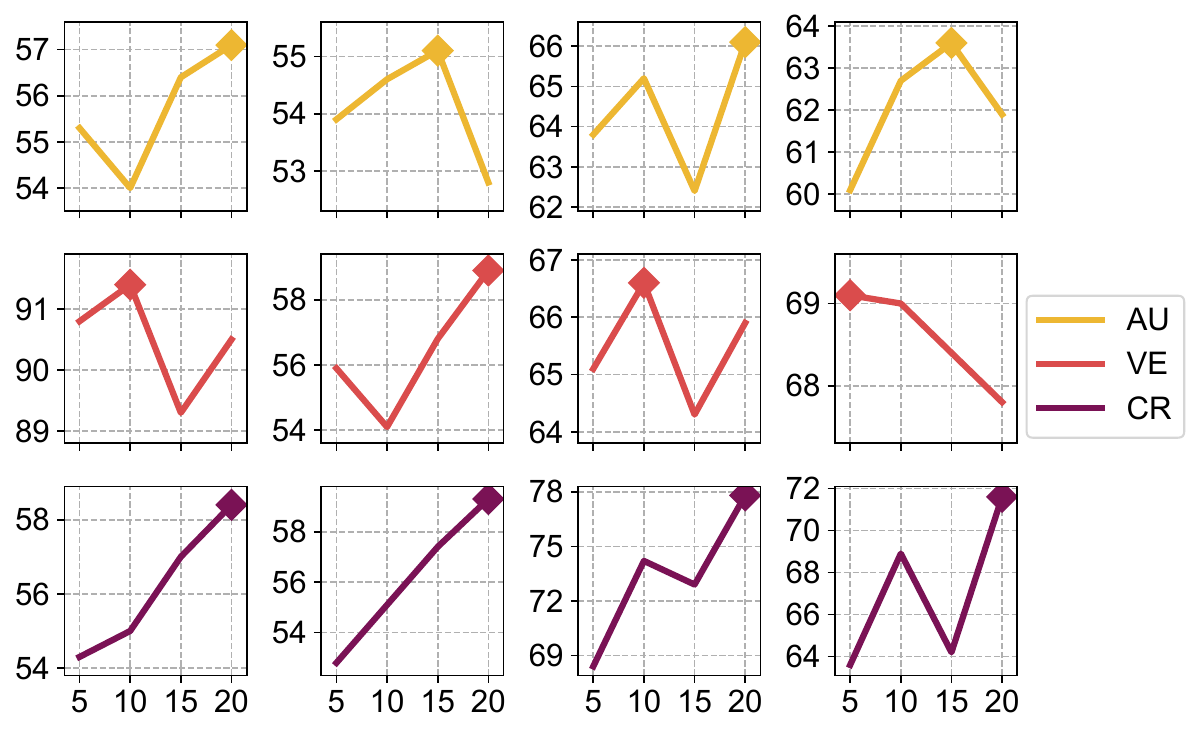}};
    
    \footnotesize
    \draw (-3.4, 3.4)  node {SR(10)};
    \draw (-3.4 + 2.3, 3.4)  node {PR(10)};
    \draw (-3.4 + 2 * 2.3, 3.4)  node {DP(20)};
    \draw (-3.4 + 3 * 2.3, 3.4)  node {PS(20)};
    \draw (0, -3.3)  node {Augmentation Rate (\%)};
    
    \node[rotate=90] at (-5, 0) {Test Accuracy (\%)};
    
    \end{tikzpicture}
    \caption{The effect of rate of our LPA augmentations on the accuracy across datasets. The diamond markers indicate the maximum achieved accuracy which was the rate that was subsequently chosen to produce the heatmap of Figure \ref{fig: heatmap}.}
    \label{fig: our_hyper}
\end{figure}

We also investigated the effect of augmentation rates for \laas and \lpas, with results presented in Figure \ref{fig: gcl_hyper} and \ref{fig: our_hyper}. 

\label{appendix: hyperparameter}

\section{Datasets}\label{appendix: dataset}

\subsection{Brief Description}
\begin{itemize}
    \item \textbf{SR \citep{selsam2018learning}} A random SAT generator proposed as a challenge for neural networks to learn intrinsic properties about satisfiability without cheating on some miscellaneous statistics about the dataset.  

    \item\textbf{PR \citep{ansotegui2009towards}} A random k-SAT generator where the frequency of each variable is sampled from a power-law distribution. 

    \item\textbf{DP \citep{ansotegui2009towards}} A pseudo-industrial generator based on PR with varying clause length

    \item\textbf{PS \citep{giraldez2017locality}} A pseudo-industrial generator based on the notion of locality. 

    \item\textbf{UR} A random k-SAT generator where each variable is sampled uniformly.
    
    \item\textbf{CA \citep{giraldez2015modularity}} A pseudo-industrial generator based on the notion of modularity.
\end{itemize}
\subsection{Parameters}
For the purpose of reproducibility, we show the parameters of each generator used in the paper below:

\begin{itemize}
    \item \textbf{SR(10) / SR(40)} All parameters follow NeuroSAT. 
    \item \textbf{PR(10)} Number of variables: 10. Number of clauses: 41. Variable per clause: 3. Power-law exponents of variables: 1.7.
    \item \textbf{PR(40)} Number of variables: 40. Number of clauses: 147. Variable per clause: 3. Power-law exponents of variables: 2.5.
    \item \textbf{DP(20)} Number of variables: 20. Number of clauses: 34. Average variables per clause: 4. Power-law exponents of variables: 1.7.
    \item \textbf{DP(40)} Number of variables: 40. Number of clauses: 75. Average variables per clause: 5. Power-law exponents of variables: 1.7.
    \item \textbf{PS(20)} Number of variables: 20. Number of clauses: 58. Min variable per clause: 2. Average variables per clause: 4.
    \item \textbf{PS(40)} Number of variables: 40. Number of clauses: 73. Min variable per clause: 2. Average variables per clause: 5.
\end{itemize}

\section{Does our method work with other GNN architectures?}
We also studied if our framework could be used with more common architectures, such as, graph convolutional networks (GCNs). We chose the GCN architecture in NeuralDiver \citep{nair2020solving} without the residual connection. The number of layers is set to be $10$. Table \ref{table: linear_gcn} shows that SSL + LPA still dominates in the low-label regime. 
\label{appendix: other_gnn}
\begin{table}[h]
\centering
\caption{Linear evaluation performance of different methods with GCNs on SR(10).}
\label{table: linear_gcn}
\vskip 0.15in
\begin{tabular}{lllllll}
\toprule
           & 2     & 10    & 100   & 1000  & 5000  & 10000 \\
\cmidrule{2-7}
Supervised & 49.82 & 50.08 & 50.68 & 50.23 & 51.25 & 78.82 \\ 
SSL + LPA  & 60.35 & 70.26 & 73.17 & 75.29 & 75.27 & 77.54 \\ 
SSL + LAA  & 51.32 & 51.29 & 48.93 & 50.72 & 50.41 & 63.24 \\
\bottomrule
\end{tabular}
\end{table}

\section{Does our method work with other contrastive loss functions?}\label{appendix: vicreg}
We replaced the SimCLR's loss function in Equation \ref{equ: contrastive} with the VICReg loss \citep{bardes2021vicreg}. We set $\lambda = 15, \mu = 1, \nu = 1$ for the hyperparameters of VICReg. As shown in Table \ref{table: vicreg}, linear evaluation performance for both objective functions are quite close. 

\begin{table}[h]
\centering
\caption{Linear evaluation performance of our methods with SimCLR and VICReg loss on SR(10)}
\label{table: vicreg}
\vskip 0.15in
\begin{tabular}{lllllll}
\toprule
              & 2     & 10    & 100   & 1000  & 5000  & 10000 \\ 
\cmidrule{2-7}
Ours + SimCLR & 79.53 & 88.32 & 92.23 & 93.32 & 93.01 & 95.12 \\ 
Ours + VICReg & 76.38 & 89.15 & 91.27 & 93.98 & 94.01 & 94.77 \\ 
\bottomrule
\end{tabular}
\end{table}

\section{More Transfer Learning Results}
\begin{figure}[h]
    \centering
    \begin{tikzpicture}
    \node at (0,0) {\includegraphics[scale=0.5]{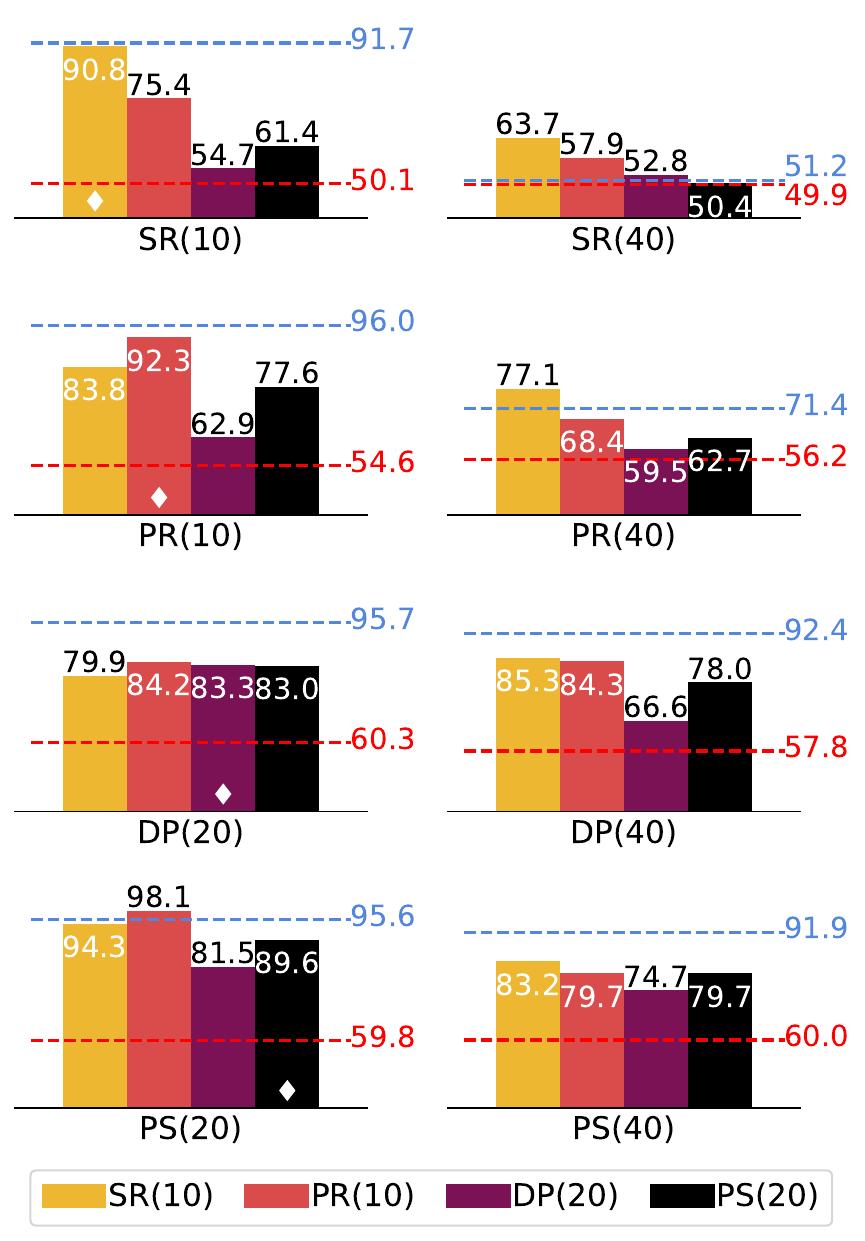}};
    \end{tikzpicture}
    \caption{ Each bar represents the SSL model trained on a different source dataset, and each subplot is the target dataset $4$ SSL models are evaluated on. The number on top of each bar is the linear evaluation accuracy with 20 labels from the target dataset. The blue line is the fully-supervised baseline trained on $20$ labels on the target baseline, and the red line is the supervised baseline trained on $10000$ labels.  The $\blacklozenge$ symbol emphasizes that the train and test datasets are from the same distribution. \emph{Left}: Transfer to unseen problems of similar size.  \emph{right}: Transfer to unseen problems of larger size.}
    \label{fig: experiment_transfer}
\end{figure}

See Figure \ref{fig: experiment_transfer}.
\label{appendix: transfer}

\section{Evaluation on Smaller Datasets}
\begin{figure*}[h]
    \centering
    
    \begin{tikzpicture}
    \node at (0,0) {\includegraphics[scale=0.41]{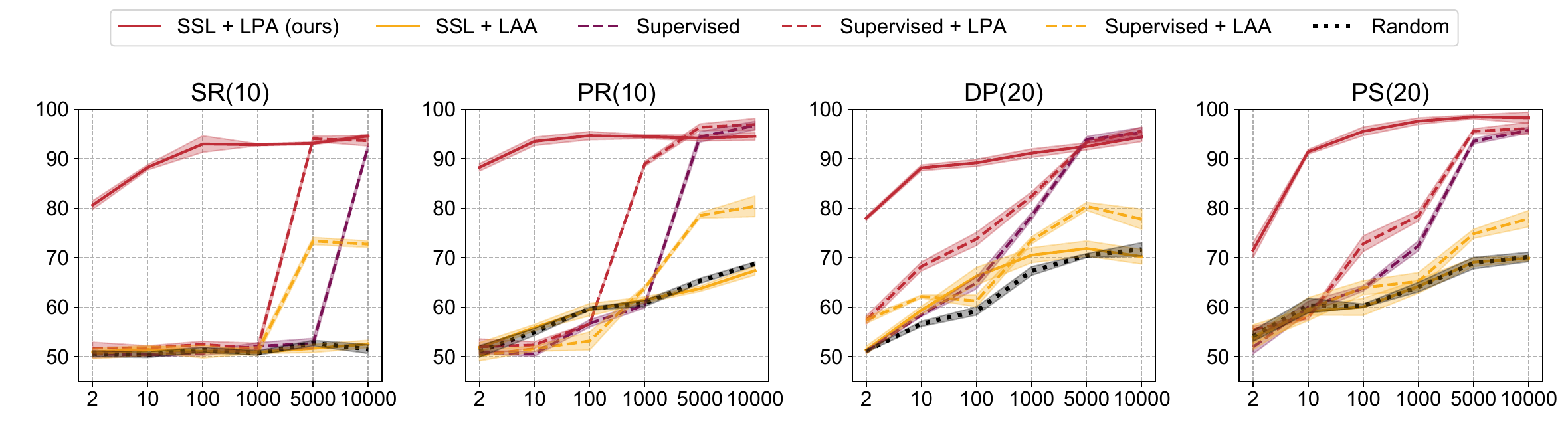}};
    \node at (0,-4) {\includegraphics[scale=0.41]{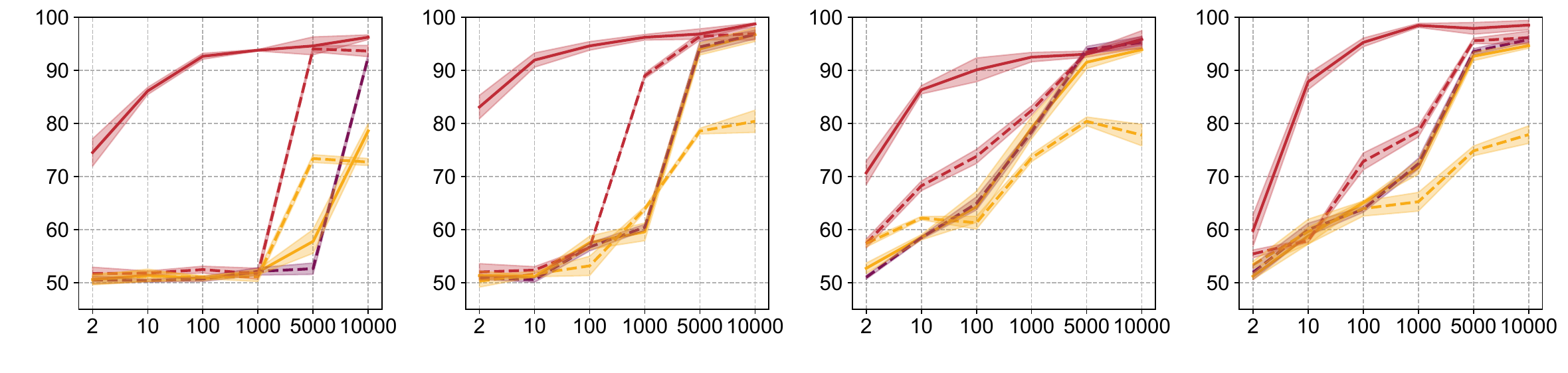}};
    \node[bag, rotate=90] at (-8.4, -0.2) {Linear Evaluation};
    \node[bag, rotate=90] at (-8.4, -3.5) {Fine-tuning};
    \scriptsize
    \node[bag, rotate=90] at (-8, -0.2) {Accuracy (\%)};
    \node[bag, rotate=90] at (-8, -3.5) {Accuracy (\%)};
    
    \draw (-5.75, -5.8)  node {No. of Training Labels};
    \draw (-1.75, -5.8)  node {No. of Training Labels};
    \draw (2.5, -5.8)  node {No. of Training Labels};
    \draw (6.5, -5.8)  node {No. of Training Labels};
    
    \end{tikzpicture}
    \caption{Our method (SSL + LPA) achieves significantly higher accuracy after linear evaluation and fine-tuning than baselines in low-label regime and is comparable to supervised models that have been given more labels. We vary the number of training labelled instances from $2$ to $10^4$ and report the average accuracy and standard error over 3 trials for all methods.}
    \label{fig: small_result}
\end{figure*}
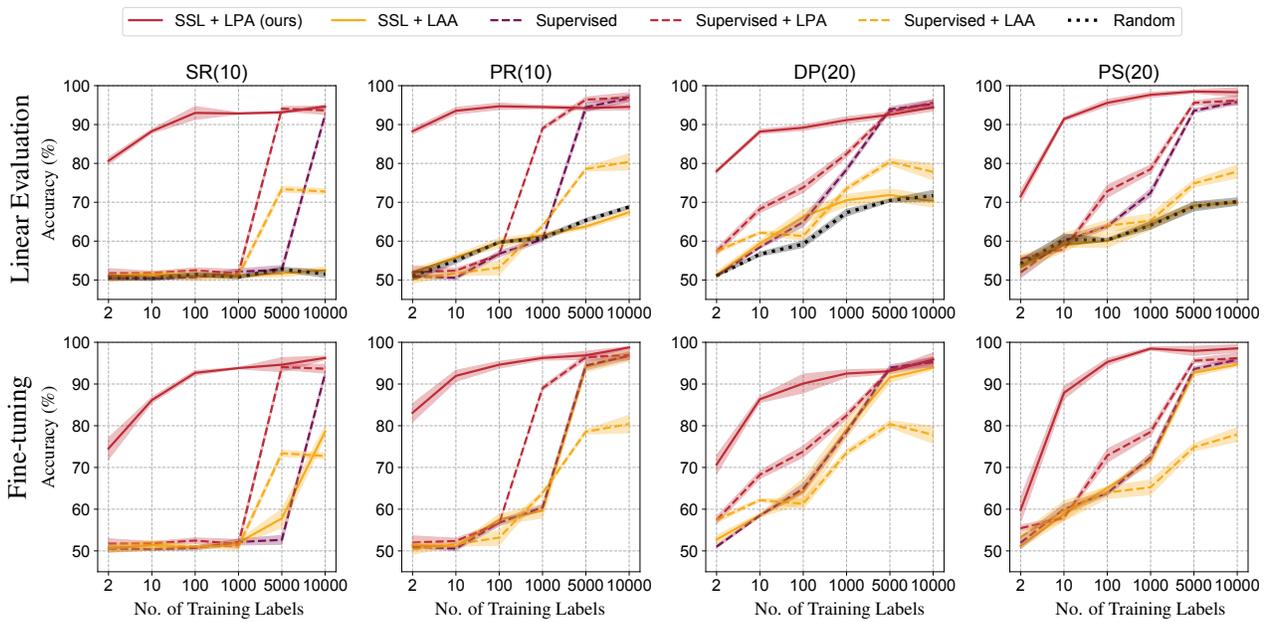

We also performed linear evaluation and fine-tuning experiments for the smaller datasets in Section \ref{sec: augmentation}. The results are shown in Figure \ref{fig: small_result}. 
\label{appendix: eval_small}

\section{Decision step} \label{appendix: decision}
We measured the decision steps of CryptoMiniSat \cite{soos2009extending} solvers before and after our augmentations. The result is shown in Table \ref{table: decision_step}. 
\begin{table}[h]
\centering
\caption{Decision steps of CryptoMiniSat \cite{soos2009extending} solvers before and after augmentations used in Figure \ref{fig: main_result}. We use the same augmentation as the corresponding SSL model in Figure \ref{fig: main_result}. The statistics is computed over $1000$ instances for each dataset. }
\label{table: decision_step}
\vskip 0.15in
\begin{tabular}{lcccccccc}
\toprule
       & \multicolumn{2}{c}{SR(40)}          & \multicolumn{2}{c}{PR(40)}          & \multicolumn{2}{c}{DP(40)}         & \multicolumn{2}{c}{PS(40)} \\      
       \cmidrule{2-9}
       & SAT & UNSAT & SAT  & UNSAT & SAT & UNSAT & SAT & UNSAT  \\ 
       \cmidrule{2-9}
      
Before & $17.54$ & $13.25 $ & $18.21$ & $12.14$ & $32.78$ & $0.27$ & $33.26$ & $7.37$ \\ 
After  & $16.36$ & $16.40$ & $21.61$ & $12.64$ & $28.47$ & $0.36$ & $34.83$ & $6.92$ \\ 
\bottomrule
\end{tabular}
\end{table}

\end{document}